\documentstyle[icip,epsfig,amssymb]{article}

\title{A Differential Invariant for Zooming}

\name{Andreas Siebert}
\address{Department of Computer Science \\
	 The University of British Columbia \\
	 siebert@cs.ubc.ca }

\begin{document}

\maketitle

\begin{abstract}
This paper presents an invariant under scaling and linear brightness change.
The invariant is based on differentials and therefore is a local feature.
Rotationally invariant 2-d differential Gaussian operators up to third order
are proposed for the implementation of the invariant.
The performance is analyzed by simulating a camera zoom-out.
\end{abstract}

\section{Introduction}

In image retrieval systems, efficiency depends prominently on identifying
features that are invariant under the transformations
that may occur, since such invariants
reduce the search space.
Such transformations typically include translation, rotation, and scaling,
as well
as linear brightness changes.
Ideally, we would like to consistently identify key points 
whose features are invariant under those transformations.

Geometrical invariants have been known for a long time,
and they have been applied more recently to vision tasks~\cite{muzi92, w93}.
Differential invariants are of particular interest since they are
local features and therefore more robust in the face of occlusion.
Building on a suggestion by Schmid and Mohr~\cite{sm97}, we propose
an invariant with respect to the four aforementioned transformations,
based on derivatives of Gaussians.

In the following, we restrict ourselves to 2-d objects, i.e.~we assume
that the objects of interest are not rotated outside the image plane,
and that they are  without significant depth
so that the whole object is in focus simultaneously.
The lighting geometry also remains constant.
In other words, we allow only translation and rotation in the image plane,
scaling that reduces the size of an object ({\em zoom-out}), and
brightness changes by a constant factor.
The zooming can be achieved by either changing the distance 
between object and camera or by changing the focal length. 

\section{The Invariant}

\subsection{The 1-d case}

Schmid and Mohr~\cite{sm97} have presented the following invariant
under scaling. 
Let $f(x)=g(u(x))=g(\alpha x)$,
i.e. g(u) is derived from f(x) by a change of variable with scaling
factor $\alpha$. 
Then $f(x)=g(u), f'(x)=\alpha g'(u), f''(x)=\alpha^2 g''(u)$,
and thus $\Theta_{12} = f'(x)^2/f''(x)$ is an invariant to scale change.
This invariant generalizes to
\begin{equation} \label{eq:SM_TheoInv}
  \Theta_{SM}(f(x)) =  \frac{[f^{(n)}(x)]^{k/n}} {f^{(k)}(x)}
\end{equation}
where $k,n \in \mathbb{N}$ denote the order of the derivatives.

$\Theta_{SM}$ is not invariant under linear brightness change.
But such an invariance is desirable because it would ensure that
properties that can be expressed in terms of the level curves
of the signal are invariant~\cite{lin94}.
A straightforward modification of $\Theta_{12}$
gives us the extended invariance:
Let $f(x) = k\,g(u) = k\,g(\alpha x)$ where $k$ is the brightness factor.
Then $f'(x)=k \alpha\,g'(u), f''(x)$ = $k \alpha^2\,g''(u),
f'''(x)=k \alpha^3\,g'''(u)$. It can be seen that
\[ \label{eq:InvScaleBright}
   \Theta_{123}(f(x)) = \frac{f'(x)\ f'''(x)}{f''(x)^2}
        = \frac{k \alpha\,g'(u)\ k \alpha^3\,g'''(u)}{(k \alpha^2\,g''(u))^2}
\]
has the desired invariance since both $\alpha$ and $k$ cancel out.
$\Theta_{123}$ can be generalized to 
$\Theta_{g3} = f^{(k)}(x)\ f^{(k+2)}(x)/$ $f^{(k+1)}(x)^2$
where $k \in \mathbb{N}$, but $k>1$ is of little interest in computer vision.

An obvious shortcoming of $\Theta_{123}$,
as well as of the other scale invariants
discussed so far, is that they are undefined where
the denominator is zero. 
Therefore, we modify $\Theta_{123}$ to be continuous everywhere:
\begin{equation}
 \label{eq:ModiInvTm123}
 \begin{tabular}{lcl}
       &0 &if c1 \\  \rule[-2.5mm]{0mm}{7mm}
    $\Theta_{m123}(f(x))=$
       &$(f'(x) f'''(x)) / f''(x)^2$
          &if c2 \\ 
       &$f''(x)^2 / (f'(x) f'''(x))$
	  &else \\
 \end{tabular}
\end{equation}
where c1 is the condition
$f''(x)=0 \wedge f'(x) f'''(x)=0$,
and c2 specifies $|f'(x) f'''(x)| < |f''(x)^2|$.
Note that this definition results in $-1 \leq \Theta_{m123} \leq 1$.

\subsection{The 2-d case}

If we are to apply eq.~\ref{eq:ModiInvTm123} to images, we have to
generalize the formula to two dimensions. Also, images are given typically
in the form of sampled intensity values rather than in the form of 
closed formulas where derivatives can be computed analytically.
One way to combine filtering with the computation of derivatives
can be provided by using Gaussian derivatives~\cite{r94, lin94, sm97}.
Let $I_1(x,y)$ and $I_2(u,v)=I_2(\alpha x,\alpha y)$ be two images related
by a scaling factor $\alpha$.
Then, according to Schmid and Mohr~\cite{sm97},
\begin{equation} \label{eq:SM2dDoGs}
 \begin{tabular}{ll}
   &$\int_{-\infty}^{\infty} I_1(x,y) \circledast
	G_{i_1\dots i_n}(x,y; \sigma)\,dx\,dy$ = \\ \rule[-2.5mm]{0mm}{8mm}
   &\ \ $\alpha^n 
   \int_{-\infty}^{\infty} I_2(u,v) \circledast
	G_{i_1\dots i_n}(u,v; \alpha \sigma)\,du\,dv$
 \end{tabular}
\end{equation}
where the $G_{i_1\dots i_n}$ are partial derivatives of the 2-d Gaussian.

Rotational invariance is a highly desirable property in most image retrieval
tasks. While derivatives are translation invariant, the partial derivatives
in~eq.~\ref{eq:SM2dDoGs} are {\em not} rotationally invariant.
However, there are some well-known rotationally invariant differential
operators. 
Recall that the 2-d zero mean Gaussian is defined as
\begin{equation} \label{eq:Gauss2dapp}
  G(x,y; \sigma)= \frac{1}{2 \pi\,\sigma^2}\ e^{-\frac{x^2+y^2}{2 \sigma^2}}
\end{equation}
Then the gradient magnitude
\begin{equation} \label{eq:Grad2d}
  |\mathrm{grad}\ G(x,y; \sigma)| 
	= \sqrt{G_x^2 + G_y^2}
	= \sqrt{x^2+y^2} / \sigma^2 \ G
\end{equation}
is a first order differential operator with the desired property.
Horn~\cite{horn86} gives the following second order operators:
\begin{equation} \label{eq:LoG}
  \begin{tabular}{ll}
   $\mathrm{LoG}(x,y; \sigma)$ 
	&= $G_{xx} + G_{yy}$ \\ 
	&= $(x^2+y^2-2\sigma^2) / \sigma^4 \ G$
 \end{tabular}
\end{equation}
\begin{equation} \label{eq:QV}
  \begin{tabular}{ll}
   & $\mathrm{QV}(x,y; \sigma)$
	 = $\sqrt{G_{xx}^2 + 2\,G_{xy}^2 + G_{yy}^2}$ \\
	\rule[-2.5mm]{0mm}{8mm}
        &\ = $\sqrt{(x^2-\sigma^2)^2 + 2 x^2 y^2 +
                                 (y^2-\sigma^2)^2} / \sigma^4 \ G$
  \end{tabular}
\end{equation}
where LoG is the Laplacian of Gaussian. QV stands for Quadratic Variation
where we have taken the square root in order to avoid high powers.
Schmid and Mohr also suggest what they call $\nu[2]$:
\begin{equation} \label{eq:MQV}
  \begin{tabular}{ll}
   & $\nu[2](x,y; \sigma)$
	= $G_{xx} G_x^2 + 2\,G_{xy} G_x G_y + G_{yy} G_y^2$ \\
	\rule[-2.5mm]{0mm}{7mm}
       &\ \ = $((x^4-\sigma^2 x^2) + 2 x^2 y^2 +
		(y^4-\sigma^2 y^2)) / \sigma^8 \ G^3$
  \end{tabular}
\end{equation}

\begin{figure}[htb]
  \epsfig{file=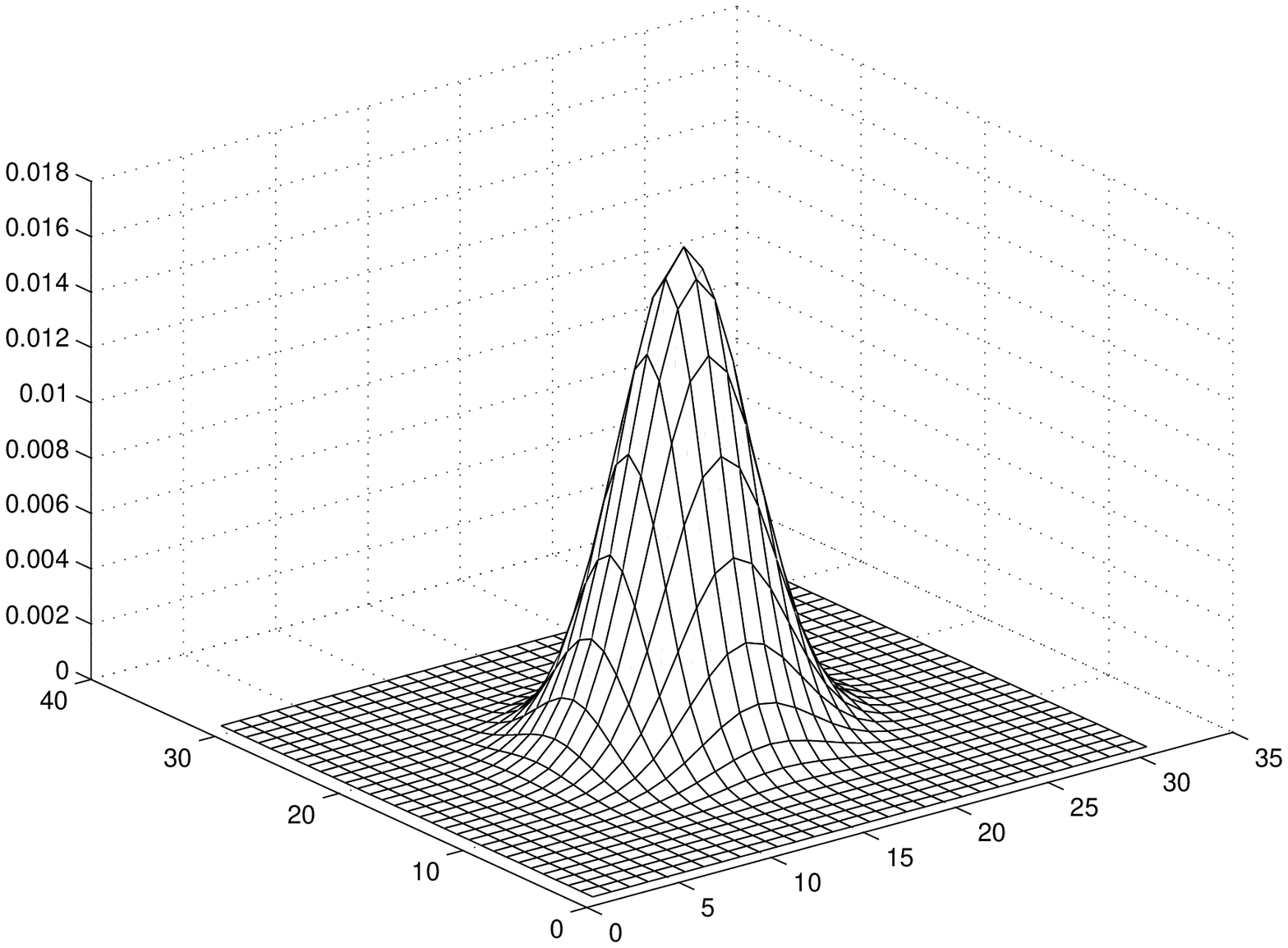,width=4.1cm,height=3.3cm}
  \epsfig{file=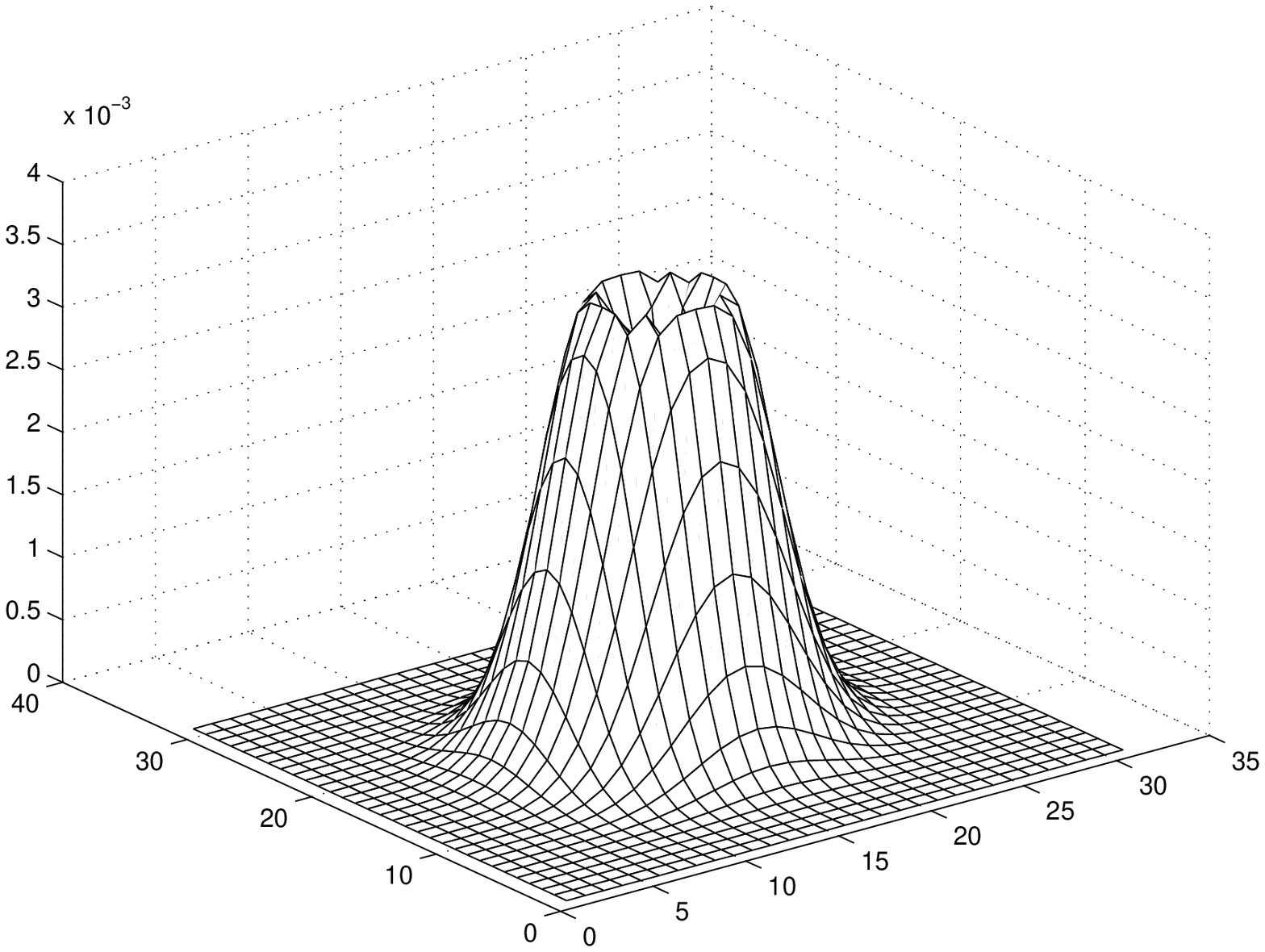,width=4.1cm,height=3.3cm}
  \epsfig{file=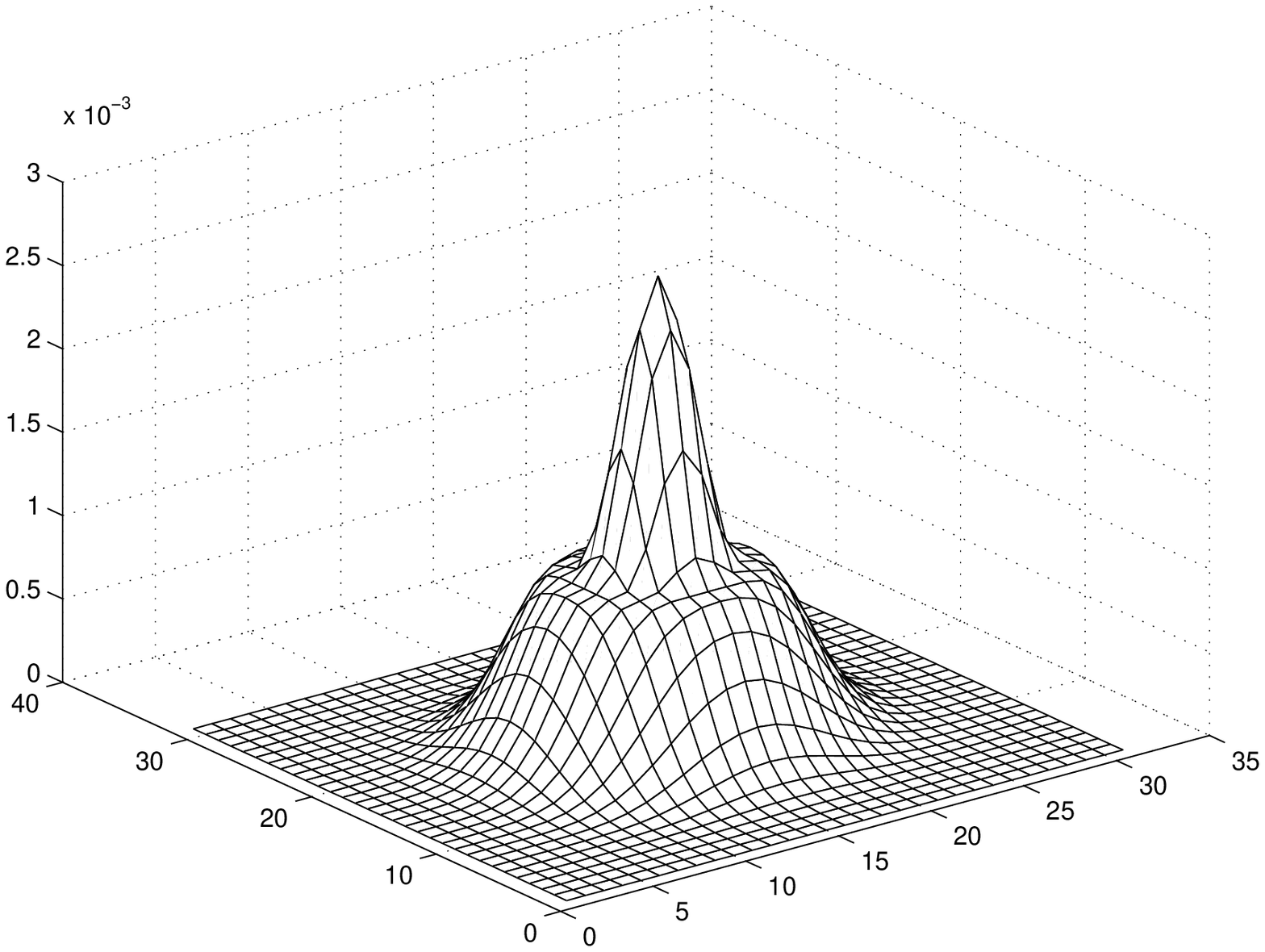,width=4.1cm,height=3.3cm}
  \epsfig{file=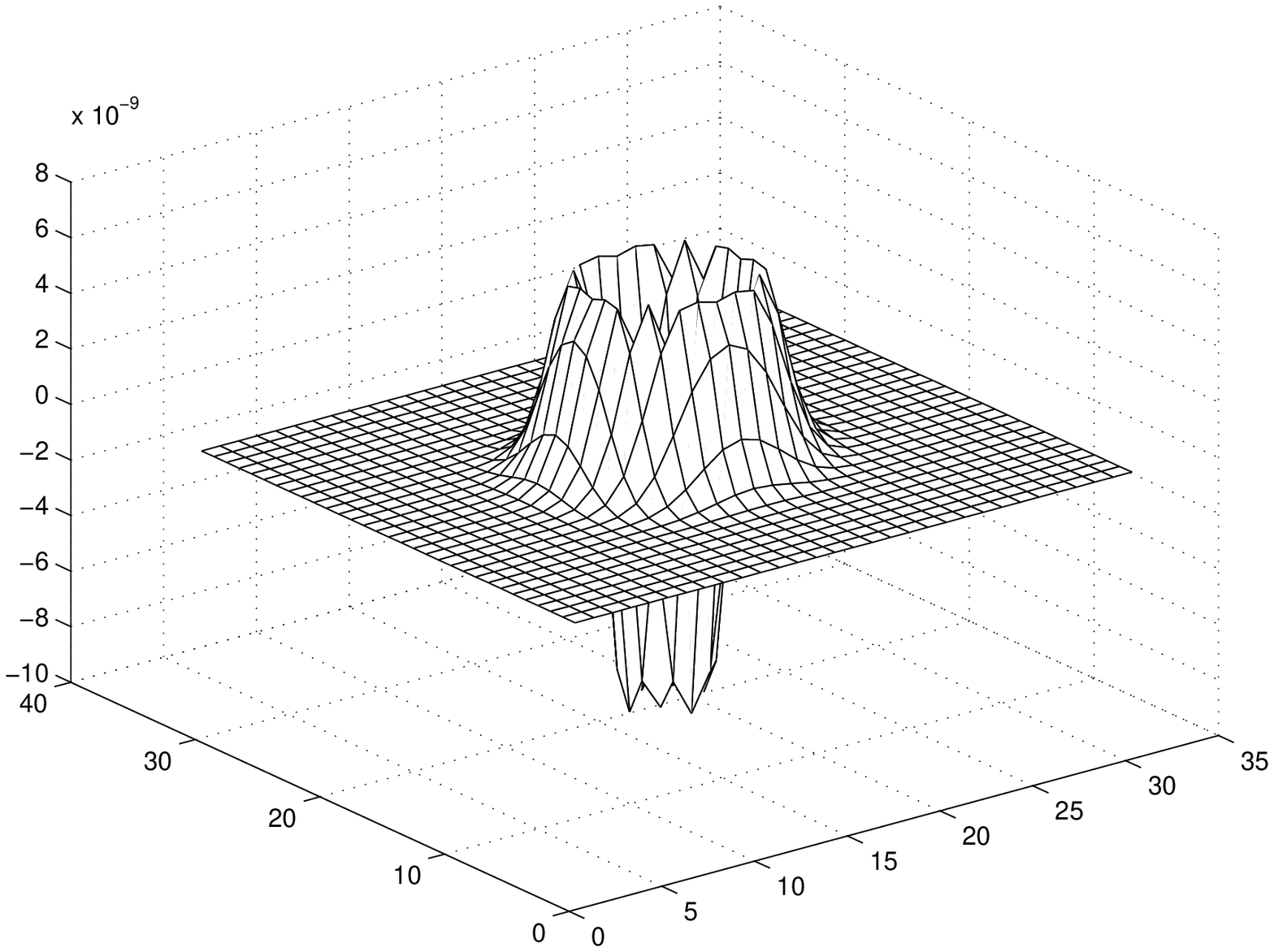,width=4.1cm,height=3.3cm}
  \epsfig{file=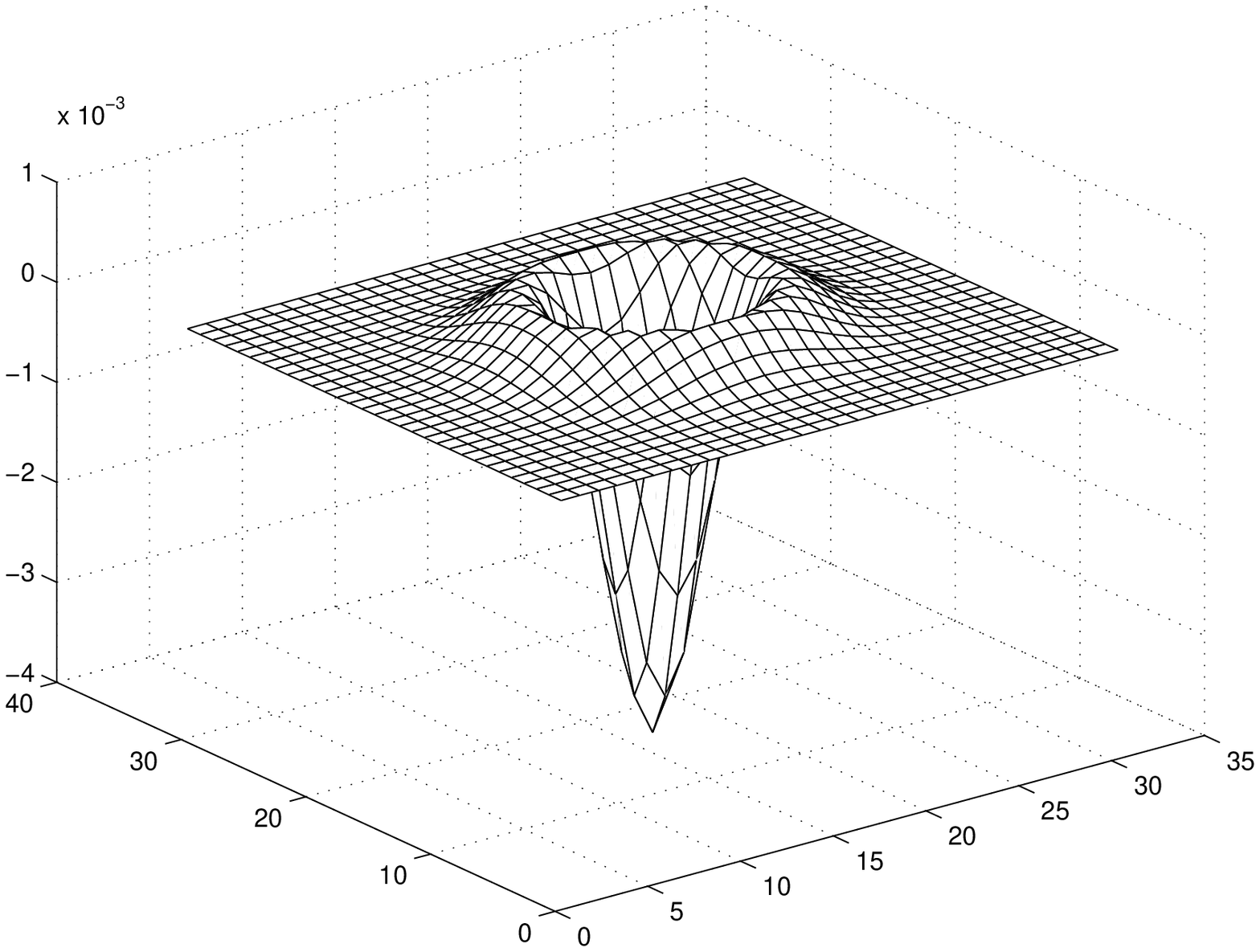,width=4.1cm,height=3.3cm}
  \epsfig{file=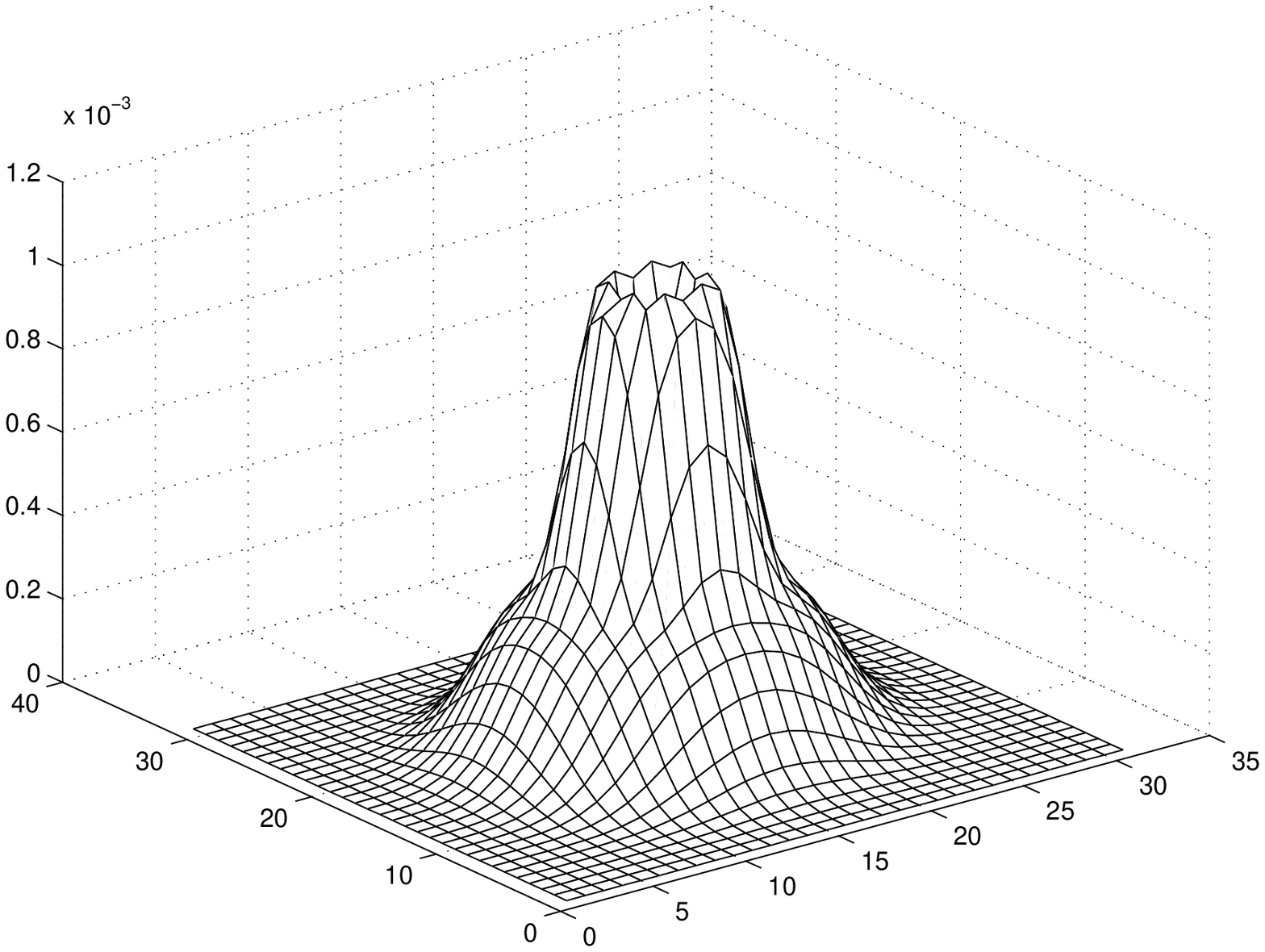,width=4.1cm,height=3.3cm}
  \epsfig{file=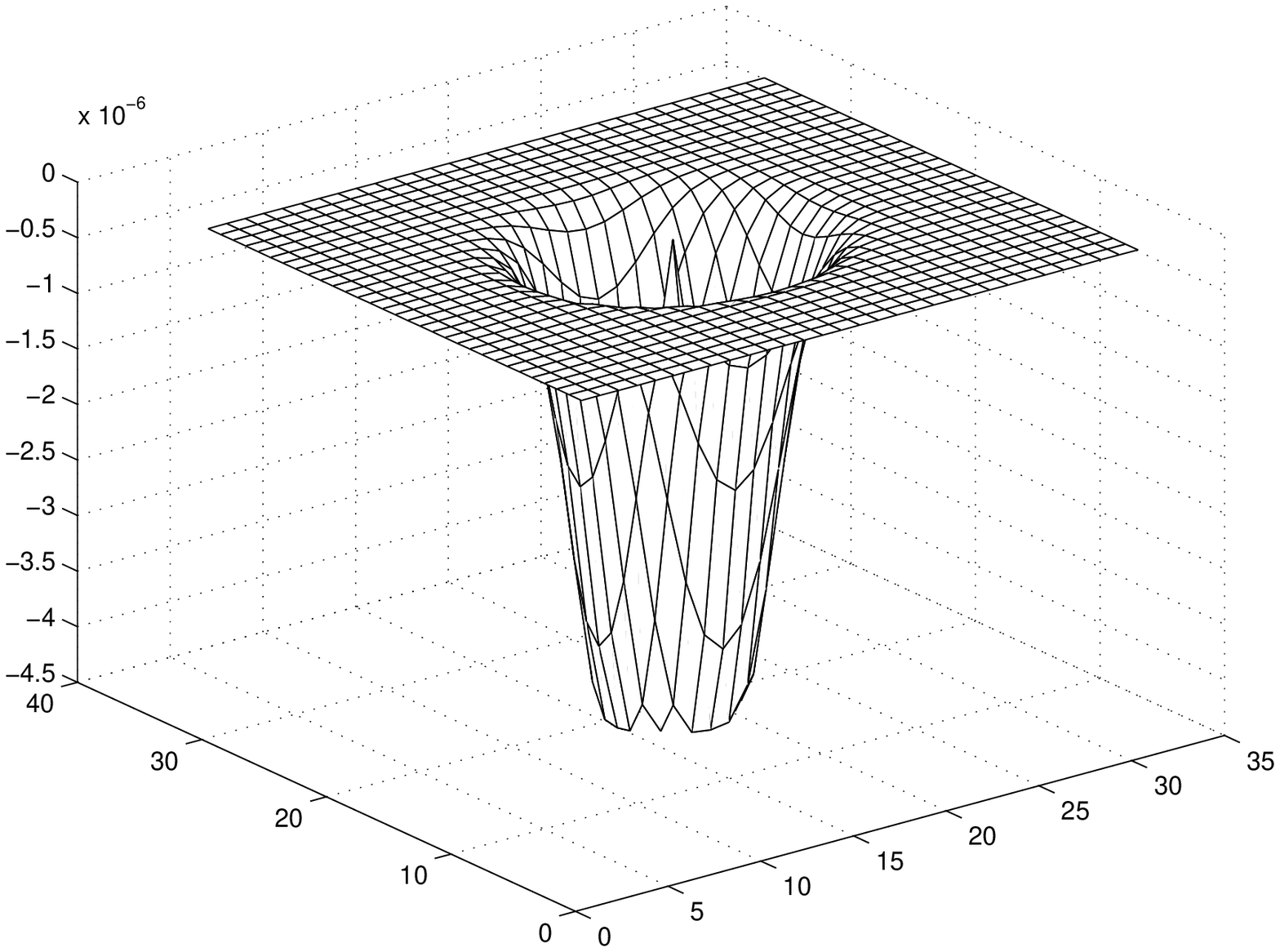,width=4.1cm,height=3.3cm}
  \epsfig{file=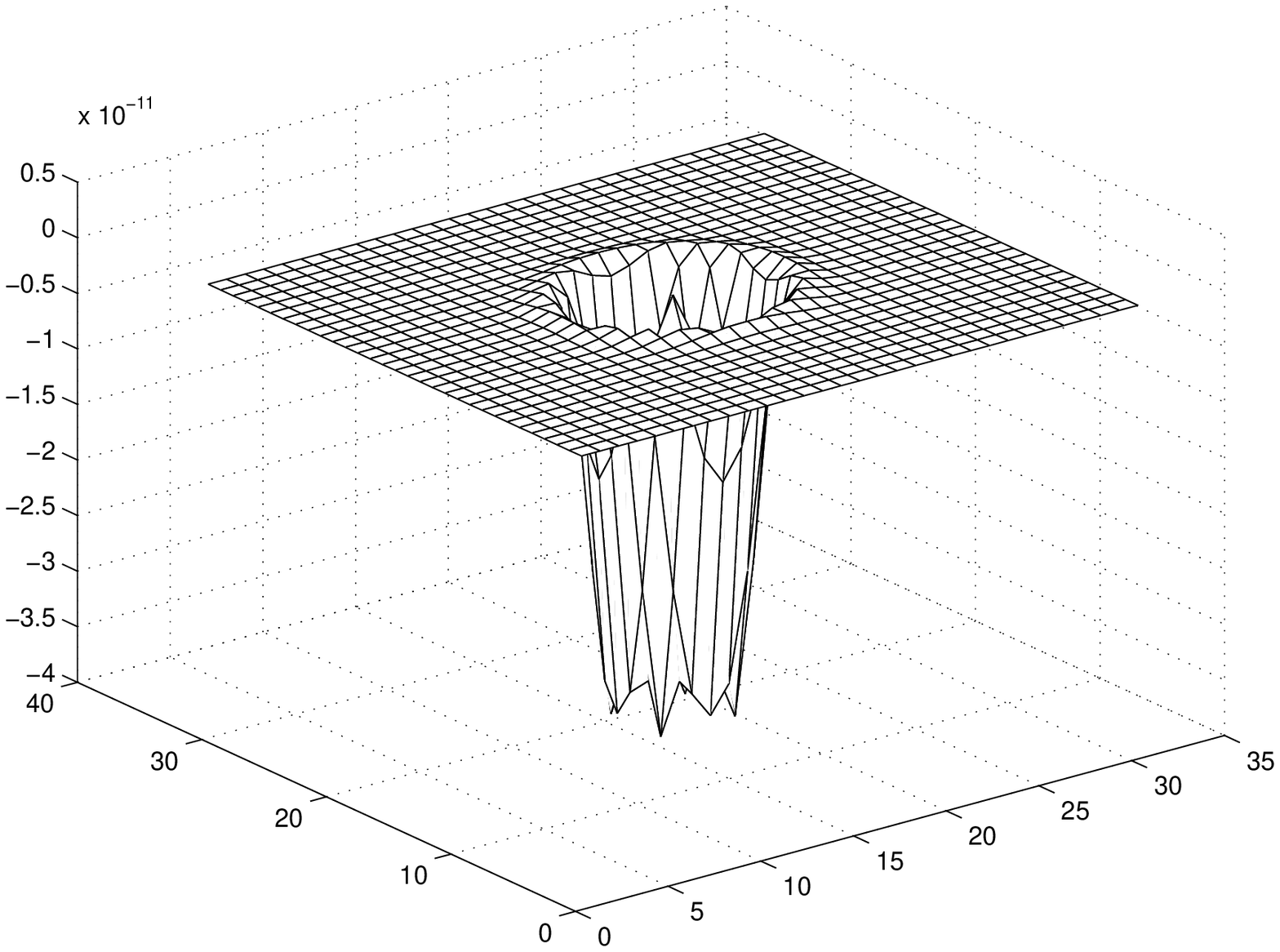,width=4.1cm,height=3.3cm}
  \caption{Rotationally invariant 2-d differential
	\mbox{operators}, $\sigma=3.0$:
	\ (a)~Gaussian \ (b)~Gradient
        \ (c)~Quadratic Variation \ (d)~$\nu[2]$
        \ (e)~Laplacian of \mbox{Gaussian} \ (f)~Cubic Variation
        \ (g)~$-\sqrt{\nu[6]}$ \ (h)~$\nu[8]$.}
  \label{fig:DiffOps}
\end{figure}

\begin{figure}[htb]
  \epsfig{file=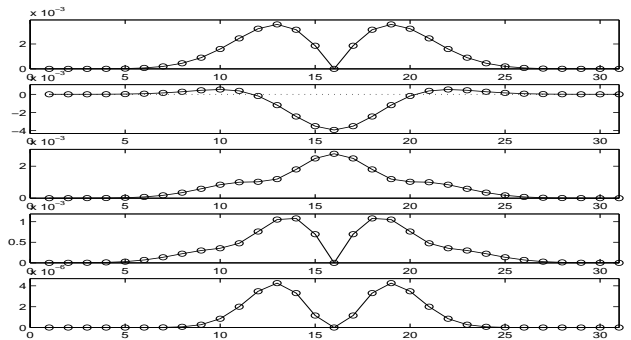,width=8.3cm,height=4.5cm}
  \caption{Cross sections through differential operators, $\sigma=3.0$:
	\ (a)~Gradient \ (b)~Laplacian of Gaussian \ (c)~Quadratic Variation
	\ (d)~Cubic Variation \ (e)~$\sqrt{\nu[6]}$.
	The circles mark the filter coefficients.}
  \label{fig:DiffOpsCross2d}
\end{figure}

Analogous to QV, we define a third order differential operator
which we call
{\em Cubic Variation} to be
\begin{equation} \label{eq:CV}
  \begin{tabular}{ll}
  & $\mathrm{CV}(x,y; \sigma)$
	 = $\sqrt{G_{xxx}^2 + 3\,G_{xxy}^2 + 3\,G_{xyy}^2 +G_{yyy}^2}$ \\
	\rule[-2.5mm]{0mm}{8mm}
	&\ \ = $\sqrt{
                        (3\sigma^2 x - x^3)^2 +
                         3 (\sigma^2 y - x^2 y)^2 + \dots} / \sigma^6 \ G$
  \end{tabular}
\end{equation}
By contrast, Schmid and Mohr use
\begin{equation} \label{eq:nu6}
  \begin{tabular}{ll}
   $\nu[6](x,y; \sigma)$
	&= $G_{xxx} G_x G_y^2 + G_{xxy} (-2\,G_x^2 G_y + G_y^3)$ \\
	 &\ \ +\ $G_{xyy} (-2\,G_x G_y^2 + G_x^3) + G_{yyy} G_x^2 G_y$ \\
	\rule[-2.5mm]{0mm}{8mm}
	&= $((3 \sigma^2 x - x^3) x y^2 + \dots) / \sigma^{12} \ G^4$
  \end{tabular}
\end{equation}
and
\begin{equation} \label{eq:nu8}
  \begin{tabular}{ll}
   $\nu[8](x,y; \sigma)$
	&= $G_{xxx} G_x^3 + 3\,G_{xxy} G_x^2 G_y$ \\
	  &\ \ +\ $3\,G_{xyy} G_x G_y^2 + G_{yyy} G_y^3$ \\
	\rule[-2.5mm]{0mm}{8mm}
	&= $((3 \sigma^2 x - x^3) x^3 + \dots) / \sigma^{12} \ G^4$
  \end{tabular}
\end{equation}
These operators are shown in fig.~\ref{fig:DiffOps} for $\sigma=3$.
Given this choice of operators, the criteria on which we select the
operators are as follows:
\begin{itemize}
  \item The operator must fulfill eq.~\ref{eq:SM2dDoGs}, i.e.~scaling
	by $\alpha$ (of both $x$ and $y$ simultaneously)
	should return a factor of $\alpha^n$, where $n$ is
	the order of the operator, so that eq.~\ref{eq:ModiInvTm123}
	is indeed a scale invariant.
  \item The operators used to compute $\Theta_{m123}$ should 
	differ in shape as much as possible from each other in order to deliver
	more discriminative results. 
\end{itemize}
With respect to the first criterion, the gradient returns a factor
$\alpha^1$, as required for a first order differential operator,
and the LoG and QV return a factor of $\alpha^2$,
but $\nu[2]$ returns $\alpha^4$. This cannot be remedied by taking the
square root since $\nu[2]$ is negative at some points.
As for the third order operators, CV returns $\alpha^3$, while
$\nu[6]$ and $\nu[8]$ return $\alpha^6$. We can take the square root
of $\nu[6]$ but not of $\nu[8]$.
Where the second criterion is concerned, the LoG is preferable to QV
since the LoG has both positive and negative coefficients, which makes it
unique compared to all other operators.
It is not obvious whether CV or $\sqrt{\nu[6]}$ has more discriminatory power,
the difference between them seems negligible.
$\sqrt{\nu[6]}$ has a slightly more compact support, but the coefficients
are an order of magnitude smaller than those of CV and the other operators.
Fig.~\ref{fig:DiffOpsCross2d} shows cross sections through the center
of some of the operators in fig.~\ref{fig:DiffOps}. 
In our experiments, we used the Gradient, LoG, and CV as the differential
operators to compute $\Theta_{m123}$ according to eq.~\ref{eq:ModiInvTm123}.
Since Gradient and CV are always positive or zero, we have
$0 \leq \Theta_{m123} \leq 1$.

Note that eqs.~\ref{eq:Grad2d} to~\ref{eq:nu8} suggest
two ways of implementation.
Either, kernels representing the partial derivatives of the Gaussian
can be used, and the operators are assembled from those kernels according to
the left hand sides of the equations, or a characteristic filter
is designed in each case
according to the right hand sides of the equations.

\section{Simulation}

\begin{figure}[htb]
  \epsfig{file=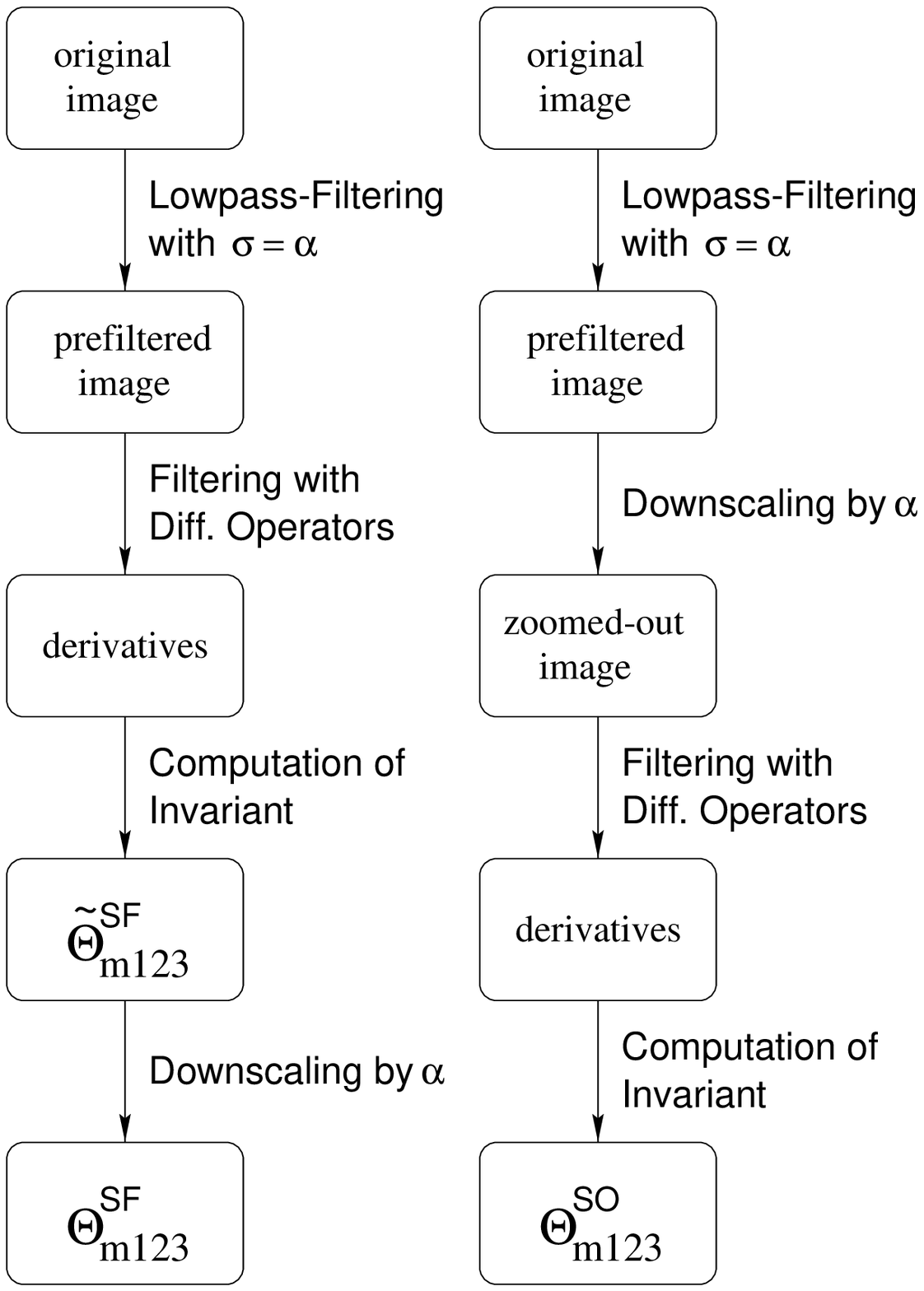,width=8.3cm,height=9.0cm}
  \caption{(left) scaling by filtering process
	vs.~(right)~scaling by simulated optical zooming.}
  \label{fig:Zoom2dSOF}
\end{figure}

The infinite integrals in eq.~\ref{eq:SM2dDoGs}
can only be approximated by sampled, finite signals and filters. 
Furthermore, in cameras, where the number of pixels is constant, a world object
is mapped into fewer pixels as the camera zooms out, leading to
increasing spatial integration over the object and ultimately to aliasing.
This means that the computation of $\Theta_{m123}$ necessarily has
an error.
Equation~\ref{eq:SM2dDoGs} suggests a way to analyze the accuracy
of $\Theta_{m123}$ by simulating the zoom-out process. 
The left hand side can be thought of as a
{\em scaling by filtering} (SF) process 
while the right hand side could be called
{\em scaling by optical zooming} (SO)
where we deal with a scaled, i.e.~reduced image
and an appropriately adjusted Gaussian operator.
Here, scaling by optical zooming serves to simulate
the imaging process as the camera moves away from an object.
The two processes are schematically depicted in fig.~\ref{fig:Zoom2dSOF}.

The input to the simulation are 8-bit images taken by a real camera.
The scaling factor $\alpha$ is a free parameter,
but it is chosen such that the downsampling maps
an integer number of pixels into integers. 
Both SF and SO start off with a lowpass filtering step. This prefiltering,
implemented by a Gaussian with $\sigma=\alpha$,
improves the robustness of the computations significantly
as derivatives are sensitive to noise.
Also, lowpass filtering reduces aliasing
at the subsequent downsampling step.

In SO, if the image function had a known analytical form,
we could do the scaling
by replacing the spatial variables $x$ and $y$ by $\alpha x$ and $\alpha y$.
But images are typically given as intensity matrices.
Therefore, the downscaling
is done by interpolation, using
cubic splines\footnote{The Matlab function \textsf{spline()} is employed}.
We then apply the differential operators (Gradient, LoG, CV)
with the appropriate value of $\alpha \sigma$ to the image
and compute the invariant $\Theta_{m123}^{SO}$.
By contrast, in SF, the operators are applied to the original size image.
The invariants are computed and then downscaled,
using again cubic spline interpolation, to the same size as
the image coming out of the SO process, so that $\Theta_{m123}^{SO}$
and $\Theta_{m123}^{SF}$ can be compared directly at each pixel.

\section{Experiments}

\begin{figure}[htb]
  \epsfig{file=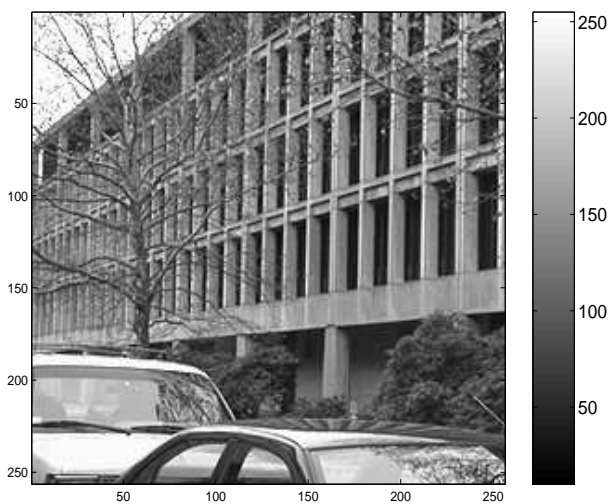,width=8.2cm,height=6.8cm}
  \epsfig{file=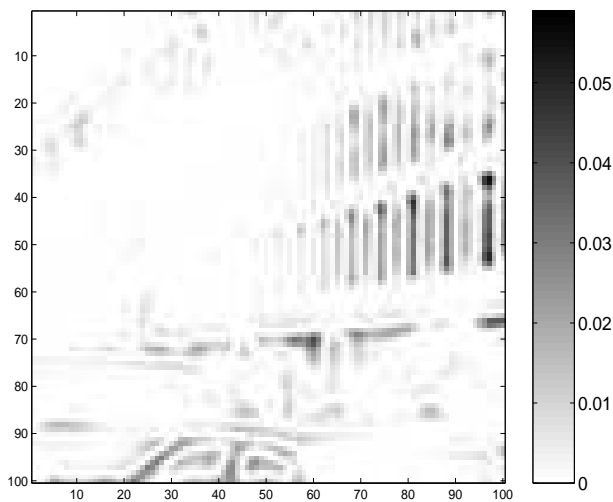,width=8.25cm,height=6.8cm}
  \epsfig{file=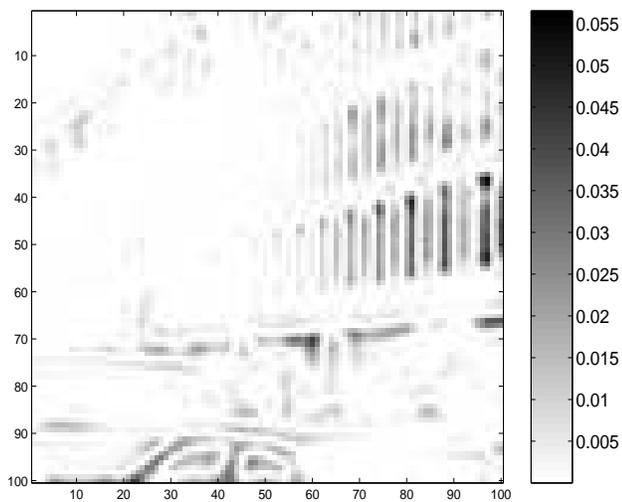,width=8.35cm,height=6.8cm}
  \caption{(a)~Original 256x256 image
	 \ (b)~$\Theta_{m123}^{SF}$ for $\alpha=2.56$
	 \ (c)~$\Theta_{m123}^{SO}$ for $\alpha=2.56$.}
  \label{fig:THETAm123_Buildings_05x256}
\end{figure}

\begin{figure}[htb]
 \begin{center}
  \epsfig{file=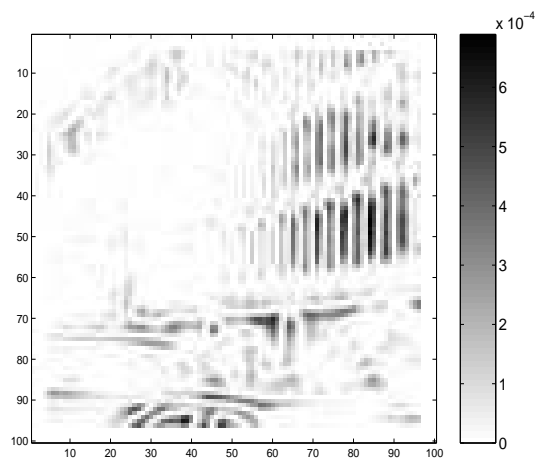,width=7.1cm,height=6.1cm}
 \end{center}
  \caption{Absolute difference $| \Theta_{m123}^{SO}-\Theta_{m123}^{SF} |$.
	Note the smaller scale ($10^{-4}$) of the error compared to 
	fig.~\ref{fig:THETAm123_Buildings_05x256}.}
  \label{fig:diffTHETAm123_Buildings_05x256}
\end{figure}

Fig.~\ref{fig:THETAm123_Buildings_05x256} demonstrates the simulation process
on a real image.
The original image, 256$\times$256, in the top row,
is downscaled to 100$\times$100,
i.e.~by a factor~$\alpha=2.56$.
The second and third row show $\Theta_{m123}$ as the results of SF and SO,
respectively, at all pixel locations.
Fig.~\ref{fig:diffTHETAm123_Buildings_05x256}
 shows the absolute difference between $\Theta_{m123}^{SF}$
and~$\Theta_{m123}^{SO}$, where the four boundary rows and columns
have been set to zero in order to mask the boundary effects.
Note that the difference is roughly a factor~100
smaller than the values of~$\Theta_{m123}^{SO}$ or~$\Theta_{m123}^{SF}$.

\begin{figure}[htb]
 \begin{center}
  \epsfig{file=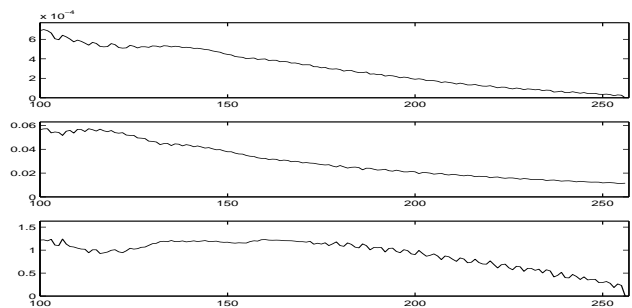,width=8.3cm,height=4.0cm}
 \end{center}
  \caption{(a) $\Delta_{\Theta_{m123}}(\alpha)$,
	 \ (b) $\max_{i,j}\, (| \Theta_{m123}^{SO} |)(\alpha)$,
	 \ (c)~$\epsilon_{gr}(\alpha)$, over image size in pixels;
	 \ $1\leq \alpha \leq 2.56$.}
  \label{fig:RelGlobErrTHETAm123_Buildings_05x256}
\end{figure}

In order to quantify the error, we have varied $\alpha$ from 1 to 2.56,
sampled such that the downscaled image has an integer pixel size,
and computed the {\em global absolute differences}
\begin{equation}
\Delta_{\Theta_{m123}}(\alpha)=
	\max_{i,j}\, (| \Theta_{m123}^{SO}-\Theta_{m123}^{SF} |)
\end{equation}
where $i,j$ range over all non-boundary pixels of the respective image,
as well as the {\em global relative error}, in percent:
\begin{equation}
\epsilon_{gr}(\alpha)=100\,\frac{\Delta_{\Theta_{m123}}}
		{\max_{i,j}\, (| \Theta_{m123}^{SO} |)}
\end{equation}
The graphs of these measures are shown
in fig.~\ref{fig:RelGlobErrTHETAm123_Buildings_05x256}.
We see that the global relative error is less than 1.3\% even as $\alpha$
becomes as large as 2.56.
For the set of images we worked with,
we have observed $0.5\% <\max (\epsilon_{gr})< 2\%$ for $\alpha<3$.
We also noted that without prefiltering $\epsilon_{gr}$ can 
become as large as 10\%.

\begin{figure}[htb]
 \begin{center}
  \epsfig{file=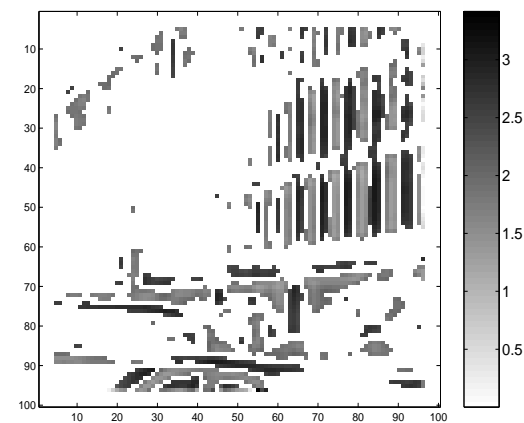,width=6.9cm,height=5.7cm}
 \end{center}
  \caption{Relative Errors at above-average values of $\Theta_{m123}^{SO}$
	for $\alpha=2.56$, in percent.}
  \label{fig:ErrorTHETAm123_Buildings_05x256}
\end{figure}

Fig.~\ref{fig:ErrorTHETAm123_Buildings_05x256}
shows the relative error per pixel in percent,
i.e.~$100\,\Delta_{\Theta_{m123}}(i,j)\,/\,| \Theta_{m123}^{SO}(i,j) |$,
but only for those pixels $(i,j)$ where $\Theta_{m123}^{SO}(i,j)$
is larger than the average
value of $\Theta_{m123}^{SO}$.
We find that error to be less than 3.5\% anywhere in the given example.

\section{Outlook}
In the context of image retrieval, there remain some
major issues to be addressed.
First, the performance of the proposed invariant has to be analyzed 
on sequences of images taken at increasing object-camera distance,
i.e.~the simulation of the zoom-out process has to be replaced with
a true camera zoom-out.
Intrinsic limitations of the image formation process
by CCD cameras~\cite{h98} can be expected
to somewhat decrease the accuracy of the invariant.

Second, a scheme for keypoint extraction must be devised.
For efficiency reasons, matching will be done on those keypoints only.
Ideally, the keypoints should be reliably identifiable, irrespective of scale.

Third, the proposed invariant must
be combined with a scale selection scheme. Note that in the simulation
above, we knew {\em a priori} the right values
for $\alpha$ and therefore for $\alpha \sigma$
and the corresponding filter size. But such is not the case in general
object retrieval tasks.
Selecting stable scales is an active research area~\cite{lin94,l99}.

\section{Acknowledgements}
The author would like to thank Bob Woodham and David Lowe for their 
valuable feedback.



\end{document}